\documentclass{isprs} 
\usepackage[labelfont=bf, labelsep=colon]{caption}
\usepackage{subfigure}
\usepackage{setspace}
\usepackage{geometry} 
\usepackage{epstopdf}
\usepackage{amsmath}

\usepackage{multirow}

\usepackage[labelsep=period]{caption}  
\usepackage[british]{babel} 
\usepackage[hang]{footmisc}
\usepackage{amssymb}

\begin{document}

\title{Temporal-Spatial Tubelet Embedding for Cloud-Robust MSI Reconstruction using MSI-SAR Fusion: A Multi-Head Self-Attention Video Vision Transformer Approach}
\date{}


\author{
 Yiqun WANG\textsuperscript{1}, Lujun LI\textsuperscript{1}, Meiru YUE\textsuperscript{1}, Radu STATE\textsuperscript{1} }

\address{
	\textsuperscript{1 }University of Luxembourg, 1359 Kirchberg, Luxembourg\\ - (yiqun.wang, lujun.li, meiru.yue, radu.state)@uni.lu\\
}



\abstract{


Cloud cover in multispectral imagery (MSI) significantly hinders early-season crop mapping by corrupting spectral information. Existing Vision Transformer(ViT)-based time-series reconstruction methods, like SMTS-ViT, often employ coarse temporal embeddings that aggregate entire sequences, causing substantial information loss and reducing reconstruction accuracy. To address these limitations, a Video Vision Transformer (ViViT)-based framework with temporal-spatial fusion embedding for MSI reconstruction in cloud-covered regions is proposed in this study. Non-overlapping tubelets are extracted via 3D convolution with constrained temporal span $(t=2)$, ensuring local temporal coherence while reducing cross-day information degradation. Both MSI-only and SAR-MSI fusion scenarios are considered during the experiments. Comprehensive experiments on 2020 Traill County data demonstrate notable performance improvements: MTS-ViViT achieves a 2.23\% reduction in MSE compared to the MTS-ViT baseline, while SMTS-ViViT achieves a 10.33\% improvement with SAR integration over the SMTS-ViT baseline. The proposed framework effectively enhances spectral reconstruction quality for robust agricultural monitoring.
}


\maketitle


\section{Introduction}\label{sec:Introduction}

Accurate early-season crop mapping is essential for agricultural monitoring, food security assessment, and sustainable resource management \cite{qader2021role}. Timely identification of crop types and phenological stages enables precision farming, optimized resource allocation, and early detection of crop stress conditions \cite{wang2023early,wang2024cross,wang2024cross2}. Multispectral imagery (MSI) from satellites such as Sentinel-2 \cite{main2017sen2cor} provides rich spectral information spanning visible to shortwave infrared wavelengths, enabling precise crop classification and continuous phenological monitoring. However, MSI observations are inherently vulnerable to cloud cover, which frequently obscures critical observations during the early growing season. Cloud-induced data gaps severely limit the ability to capture temporal crop dynamics and make informed agricultural decisions, a problem particularly acute during early growth stages when phenological changes are most indicative of crop variety and health status.

Synthetic aperture radar (SAR) data from sensors such as Sentinel-1 \cite{torres2012gmes} provides a complementary solution, as microwave signals penetrate cloud cover and deliver consistent all-weather, day-and-night observations. However, SAR fundamentally lacks the spectral richness and spatial detail of MSI, rendering it insufficient for precise discrimination between crop types at the spatial and spectral granularity required for operational agricultural applications. Existing approaches to address cloud cover rely on simplistic techniques such as linear temporal interpolation \cite{xia2012estimation,wang2024cross,wang2024cross2}, closest spectral fit \cite{eckardt2013removal}, or single-timeframe MSI-SAR fusion \cite{scarpa2018cnn,li2019thick,jing2023denoising,10341330,tu2025cloud}, which fail to capture the complex temporal-spectral patterns necessary for accurate reconstruction under severe and prolonged cloud cover. A recent advanced multi-head self-attention Vision Transformer(ViT)-based model called SMTS-ViT \cite{li2025vision} for time-series MSI reconstruction leverages integrated MSI-SAR fusion to achieve superior results compared to single-time frameworks. However, it employs temporal embeddings that aggregate information uniformly across the entire sequence, which can lead to significant information loss due to temporal aliasing and ultimately degrade reconstruction quality.


In this paper, a Video Vision Transformer \cite{arnab2021vivit} (ViViT)-based framework is proposed, which introduces temporal-spatial fusion embedding via 3D convolutional tubelet projection with constrained temporal span, rather than aggregating information uniformly across the entire time series. This thereby preserves local temporal coherence and reduces information loss from excessive temporal aggregation. Theoretically, the original convolutional patch projection from \cite{li2025vision} can be understood as a degenerate case of our time-spatial fusion embedding where temporal span equals the full sequence length. By limiting the temporal span, our approach reduces cross-day information degradation and more effectively captures transient spectral patterns essential for accurate MSI reconstruction in cloud-covered regions.

In conclusion, three main contributions are presented in this study: (1) A novel temporal-spatial fusion embedding method is introduced, enabling robust multispectral cloud removal and reconstruction through explicit integration of MSI and SAR data. (2) The masking strategy for SAR inputs is improved. Unlike the previous method \cite{li2025vision} that masks SAR whenever clouds are present in MSI, SAR data remain unmasked and available, so its all-weather sensing properties are fully leveraged even in cloud-affected scenes. (3) The effectiveness of SAR-MSI fusion and the proposed temporal-spatial embedding framework, compared with MSI-only and standard ViT-based approaches, has been comprehensively validated by a carefully designed experimental setup and quantitative evaluation, clearly demonstrating superior performance under various cloud conditions.


\section{Related Works}\label{RelatedWorks}

Cloud cover remains a pervasive obstacle in optical remote sensing for agricultural monitoring, often obscuring important ground information needed for accurate analysis. Traditional gap-filling strategies, such as temporal interpolation \cite{xia2012estimation,wang2024cross,wang2024cross2} or closest spectral fit \cite{eckardt2013removal}, are limited in their ability to represent the underlying spectral and spatial variability crucial for monitoring crop development. Although advanced statistical inpainting methods with spatial regularization have been proposed, their effectiveness diminishes when large, contiguous cloud-covered regions occur during key phenological stages, leaving substantial gaps in usable observations.

Recent developments have shifted toward leveraging data fusion and deep learning to address these limitations. Notably, integrating SAR with optical MSI enables all-weather observation and supports robust cloud removal through multi-source data fusion, as demonstrated by deep residual neural networks and spatial-spectral network architectures. Approaches such as Convolutional Neural Network(CNN)-based networks \cite{scarpa2018cnn,li2019thick}, denoising diffusion models \cite{jing2023denoising}, and multi-modal similarity attention mechanisms \cite{tu2025cloud} have delivered significant improvements in Sentinel-2 imagery reconstruction by explicitly modeling the dependence between SAR and MSI inputs. While these multi-source fusion methods show clear advantages over single-modal and traditional approaches, these methods typically aggregate or process multi-modal data without explicitly leveraging time-series information, treating each image independently or focusing solely on spatial relationships.

Vision Transformers (ViTs) \cite{vaswani2017attention}, with their powerful multi-head self-attention mechanism, have shown remarkable advantages in remote sensing image reconstruction. Unlike traditional CNNs that are limited by fixed local receptive fields, the multi-head self-attention in ViTs can efficiently capture both global and local dependencies across spatial and temporal dimensions, leading to richer feature representations and improved reconstruction quality. This global modeling ability allows ViTs to better leverage important temporal and spatial context, making them particularly effective for handling cloud-contaminated multispectral series.

Recent advanced work by \cite{zhao2023seeing} proposes the Space-Eye network, which utilizes ViTs with single-head attention for cloud removal by processing time-series SAR and MSI data. This approach represents an early fusion of multi-modal and multi-temporal data, but the attention mechanism is limited in modeling complex dependencies due to the use of only a single attention head.

Nevertheless, \cite{li2025vision} introduces the SMTS-ViT architecture, which employs ViTs with multi-head self-attention to more effectively exploit temporal and spatial coherence in MSI and SAR data for cloud-filling applications. By leveraging multi-head self-attention, SMTS-ViT can model richer and more diverse relationships across multiple dates and modalities, leading to improved reconstruction quality in cloud-covered regions. This marks a shift towards explicit time-series modeling and highlights the value of temporal aggregation and advanced attention mechanisms for robust agricultural monitoring under clouds.

However, SMTS-ViT aggregates temporal information across the entire sequence, which can dilute local spectral variations critical for precise cloud gap-filling. To address this, we propose SMTS-ViViT, an enhanced ViViT-based framework that leverages temporal-spatial fusion embedding with a controlled temporal span using 3D convolutional tubelet projection. Unlike previous approaches that over-aggregate temporally or rely on computationally intensive separate spatial and temporal embeddings, our method extracts nonoverlapping spatio-temporal tubelets with constrained temporal span ($t=2$), thus preserving local temporal coherence and reducing information loss. Theoretically, standard convolutional patch projection from SMTS-ViT is a special case of our framework where the temporal span covers the full sequence, but limiting the span empirically improves reconstruction for both MSI-only and SAR-MSI fusion. Our comprehensive evaluation also demonstrates performance gains, establishing time-spatial fusion embedding as an effective solution for cloud-robust remote sensing image reconstruction.

\section{Study Areas and Data Preparation}\label{SA}

\begin{figure}[ht!]
\begin{center}
		\includegraphics[width=1.0\columnwidth]{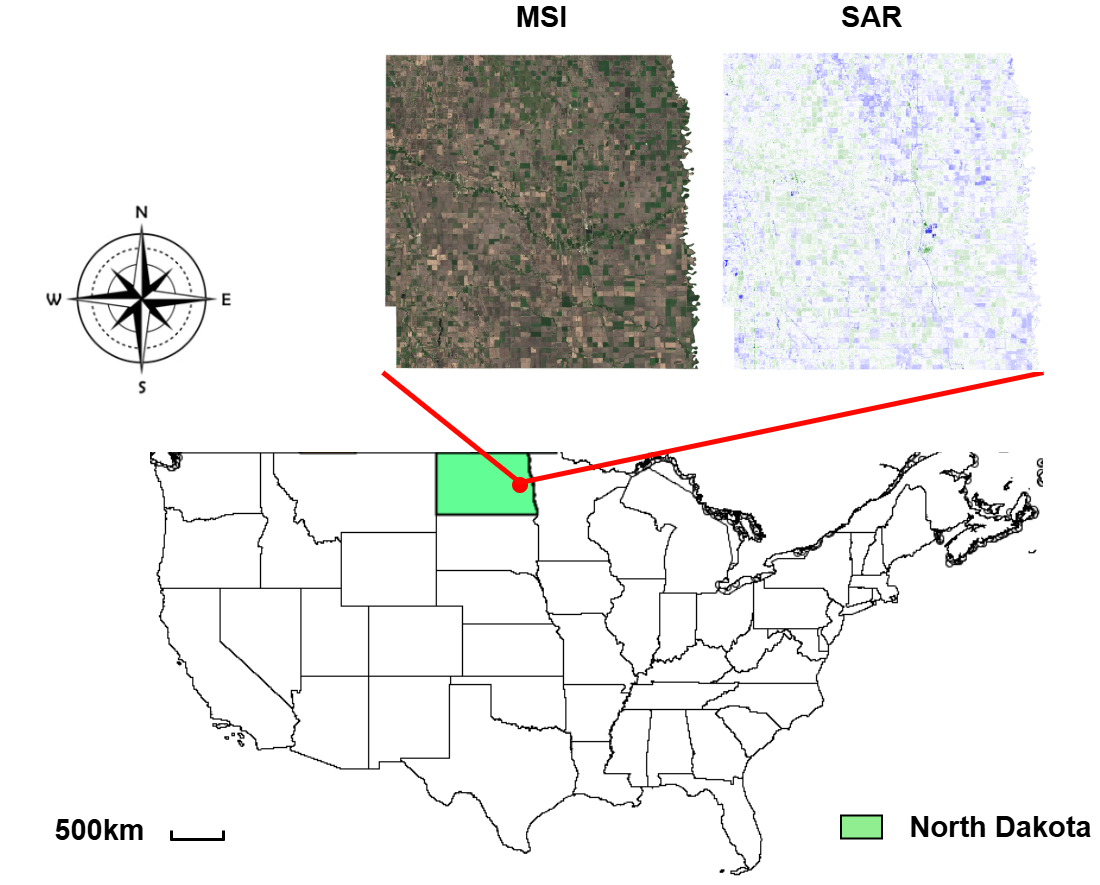}
	\caption{The Study Area: Traill County, located in North Dakota, the USA, 2020.}
\label{fig:sa}
\end{center}
\end{figure}

\subsection{Study Areas}

The study area selected for this research is Traill County, located in North Dakota, USA, as shown in Figure \ref{fig:sa}, which is an agriculturally significant region characterized by intensive crop production systems, making it an ideal testbed for evaluating time-series MSI image reconstruction and crop mapping applications. Traill County experiences pronounced seasonal variations in crop development, with early May marking the onset of critical phenological stages during corn and soybean emergence. The region's agricultural intensity and well-documented crop management practices provide a representative context for testing reconstruction methodologies under realistic cloud cover conditions.


\subsection{Data Preparation}

For this research, two complementary optical and microwave remote sensing platforms are utilized: Sentinel-1 SAR and Sentinel-2 MSI. The Sentinel-1 mission provides a consistent all-weather monitoring capability with two operational satellites maintaining a 6-day nominal revisit cycle over mid-latitude regions. The SAR data includes two polarization bands: VV (vertical transmit, vertical receive) and VH (vertical transmit, horizontal receive), which capture distinct geometric and moisture-related surface scattering properties valuable for vegetation characterization and structural analysis.

The Sentinel-2 constellation comprises two identical satellites with a combined 5-day nominal revisit frequency, delivering high-resolution multispectral observations across 13 spectral bands spanning the visible, near-infrared, and shortwave infrared regions. For this study, 11 spectral bands are utilized. They are selected for their relevance to vegetation monitoring and crop phenology analysis: B1 (Coastal Aerosol, 60 m), B2 (Blue, 10 m), B3 (Green, 10 m), B4 (Red, 10 m), B5 (Red Edge 1, 20 m), B6 (Red Edge 2, 20 m), B7 (Red Edge 3, 20 m), B8 (Near Infrared, NIR, 10 m), B8A (Red Edge 4, 20 m), B11 (Shortwave Infrared 1, SWIR1, 20 m), and B12 (Shortwave Infrared 2, SWIR2, 20 m). Additionally, cloud detection is carried out using Sentinel Hub's tools \cite{skakun2022cloud}, which generate cloud masks by filtering out cloud-covered regions in the MSI data. 

\begin{figure}[ht!]
\begin{center}
		\includegraphics[width=1.0\columnwidth]{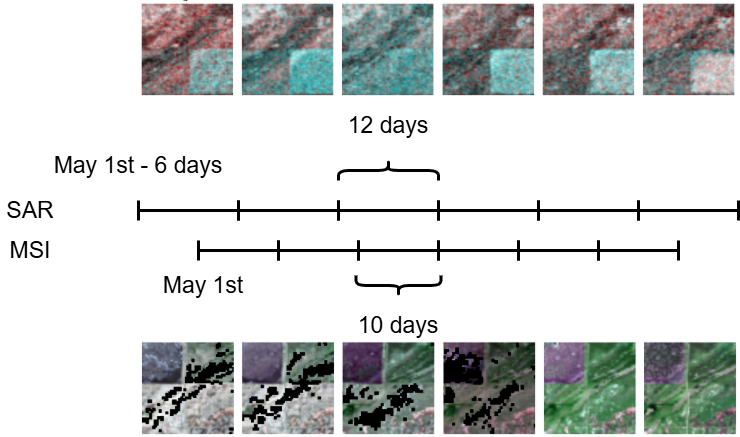}
	\caption{Multi-Temporal SAR and MSI Acquisition Scheme and Data Example. The black pixels in MSI present the cloud mask.}
\label{fig:datacollection}
\end{center}
\end{figure}

\subsubsection{Temporal Coverage and Data Collection Period:}

This paragraph details the overall temporal span and the scientific motivation for the data collection period used in the study. Our objective was to capture the critical phases of early-season crop development when phenological variability is highest and thus most informative for crop differentiation and monitoring. As shown in Figure \ref{fig:datacollection}, Sentinel-2 MSI composites were generated at 10-day intervals over a continuous 60-day window starting May 1st, covering phenomena like seed germination and early vegetative growth. For Sentinel-1 SAR, six observations were obtained by commencing acquisition six days earlier (April 25th) and extending the period to 72 days, with composited images produced every 12 days. This temporal design, shown in Figure \ref{fig:datacollection},  ensures that both MSI and SAR sequences are consistently paired and aligned with key agronomic events and realistic cloud cover variability.

\subsubsection{Spatial Resolution Harmonization and Preprocessing:}

To ensure consistent spatial resolution and geographic alignment across all data sources, we implemented comprehensive preprocessing and reprojection procedures. All Sentinel-1 SAR and Sentinel-2 MSI images were reprojected to a common Universal Transverse Mercator (UTM) projection and resampled to a uniform 30-meter ground pixel size using bilinear interpolation for spectral bands and nearest-neighbor resampling for categorical data. 


\subsubsection{Data Aggregation Strategy and Temporal Gap Management:}

There is a practical challenge posed by missing or incomplete image coverage despite Sentinel-1 and Sentinel-2’s nominal revisit cycles of 6 and 5 days, respectively. In real-world scenarios, acquisition gaps caused by satellite transmission, maintenance, processing delays, or scheduling priorities can disrupt temporal continuity. To guarantee complete and reliable time series for deep learning, we implemented a data aggregation approach: for each sensor, two successive revisit periods were composited, yielding aggregate SAR and MSI images at 12-day and 10-day intervals, respectively. This protocol ensures uninterrupted sequences, six MSI and six SAR frames as shown in Figure \ref{fig:datacollection}, across the study period. Pixelwise max compositing was employed to preserve spatial detail and minimize the effects of atmospheric interference or sensor artifacts. As a result, the final dataset maintains both sufficient temporal resolution to capture crop growth dynamics and full spatiotemporal coverage, ensuring its suitability for robust time-series modeling and fusion.

\begin{figure*}[ht!]
\begin{center}
		\includegraphics[width=2.1\columnwidth]{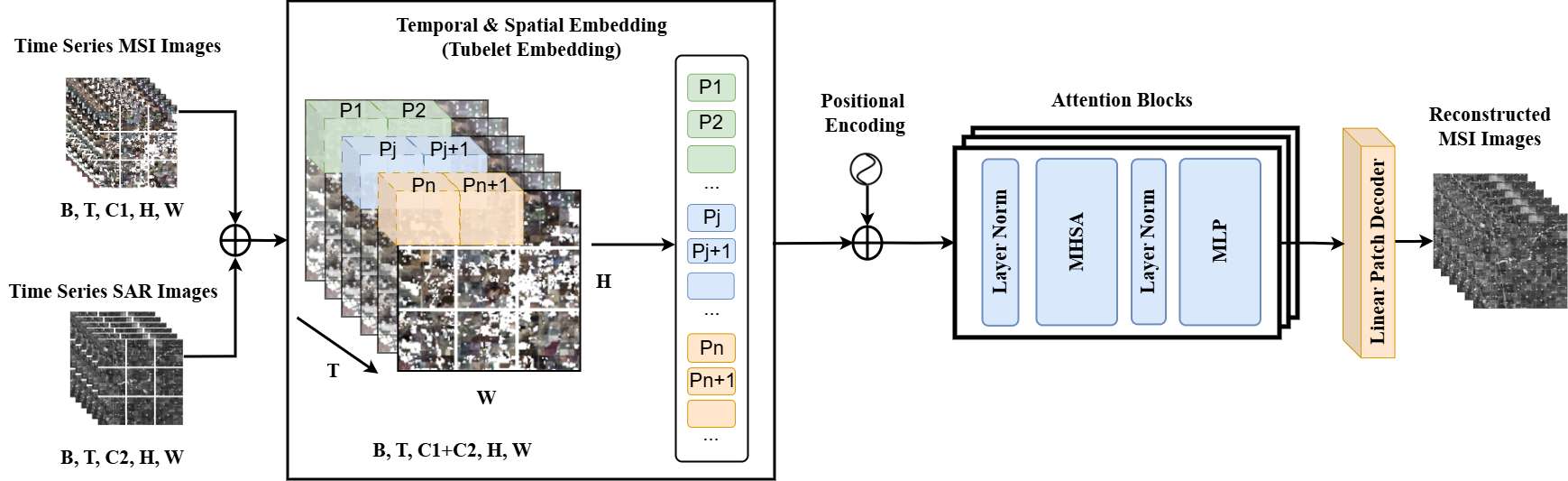}
	\caption{The Video Vision Transformer Structure with the Temporal-Spatial Tubelet Embedding for Cloud-Robust MSI Reconstruction using MSI-SAR Fusion.}
\label{fig:tubelet}
\end{center}
\end{figure*}

\section{Methodology}\label{Methodology}

\subsection{Problem Definition}\label{sec:ProblemDef}

The time-series multispectral imagery affected by cloud occlusion can be represented as $X_{\text{MSI}} \in \mathbb{R}^{T \times C_{\text{MSI}} \times H \times W}$, where $T$ denotes the temporal sequence length, $C_{\text{MSI}}$ represents the total number of spectral channels in MSI data, and $H$, $W$ denote the spatial height and width dimensions. The pixel value at position $(x, y)$ in channel $c$ at time $t$ is denoted as $X_{\text{MSI}}(t, c, x, y)$. A binary cloud mask tensor $M_{\text{cloud}} \in \mathbb{R}^{T \times H \times W}$ identifies cloud-affected pixels in the MSI observations:

\begin{equation}\label{equ:CloudMask}
M_{\text{cloud}}(t, x, y) =
\begin{cases}
0 & \text{if } X_{\text{MSI}}(t, :, x, y) \text{ is cloud-free} \\
1 & \text{if } X_{\text{MSI}}(t, :, x, y) \text{ is cloud-occluded}
\end{cases}
\end{equation}

\begin{tabbing}
where \hspace{0.8cm} \= $T$ = temporal sequence length\\
\> $C_{\text{MSI}}$ = total number of MSI spectral channels\\
\> $H, W$ = spatial dimensions\\
\> $X_{\text{MSI}}(t, c, x, y)$ = original MSI spectral value
\end{tabbing}

Assuming uniform cloud coverage across all spectral channels at each spatio-temporal location, the cloud mask $M_{\text{cloud}}$ is obtained from the Sentinel Hub's tools. Critically, the cloud mask is applied only to MSI data, as synthetic aperture radar (SAR) observations are unaffected by cloud cover and require no masking. Accordingly, SAR time-series data are represented as $X_{\text{SAR}} \in \mathbb{R}^{T \times C_{\text{SAR}} \times H \times W}$, where $C_{\text{SAR}} = 2$ (VV and VH polarizations for Sentinel-1), and all SAR observations are preserved without masking:

\begin{equation}\label{equ:SAR_Data}
X_{\text{SAR}}(t, c, x, y) \text{ for all } t, c, x, y \text{ (no masking applied)}
\end{equation}

For the MSI-only reconstruction task, the clouded MSI data $X_{\text{MSI}}^{\text{cloud}}$ is obtained by applying the cloud mask via element-wise multiplication:

\begin{equation}\label{equ:CloudedMSI}
X_{\text{MSI}}^{\text{cloud}}(t, c, x, y) = X_{\text{MSI}}(t, c, x, y) \cdot (1 - M_{\text{cloud}}(t, x, y))
\end{equation}

The ground-truth target sequence $Y \in \mathbb{R}^{T \times C_{\text{MSI}} \times H \times W}$ represents the cloud-free reference MSI imagery. For the MSI-only reconstruction scenario, the model prediction is defined as:

\begin{equation}\label{equ:Prediction_MSI}
\hat{Y}_{\text{MSI}} = \text{Model}_{\text{MSI}}(X_{\text{MSI}}^{\text{cloud}}, M_{\text{cloud}})
\end{equation}

For the SAR-MSI fusion scenario, the combined input integrates both data modalities, where MSI is cloud-corrupted but SAR remains complete and unmasked. The fused model prediction becomes:

\begin{equation}\label{equ:Prediction_Fusion}
\hat{Y}_{\text{MSI}} = \text{Model}_{\text{Fusion}}(X_{\text{MSI}}^{\text{cloud}}, X_{\text{SAR}}, M_{\text{cloud}})
\end{equation}

\begin{tabbing}
where \hspace{0.8cm} \= $X_{\text{MSI}}^{\text{cloud}}$ = cloud-corrupted MSI observations\\
\> $X_{\text{SAR}}$ = complete SAR observations\\
\> $M_{\text{cloud}}$ = MSI cloud mask\\
\> $\hat{Y}_{\text{MSI}}$ = reconstructed cloud-free MSI
\end{tabbing}

The optimization objective for both scenarios is to minimize the reconstruction loss function $\mathcal{L}(\hat{Y}_{\text{MSI}}, Y)$, ensuring that predicted values closely approximate the true spectral distribution of cloud-free MSI observations while minimizing information loss across temporal and spectral dimensions. SAR serves as a complementary data source providing all-weather context without being subject to cloud masking, enabling improved reconstruction through multi-modal information fusion.

\subsection{General Framework Architecture:} 

The proposed framework, as shown in Figure \ref{fig:tubelet} consists of three principal components: 

(1) \textbf{Temporal-Spatial Fusion Embedding} via 3D convolutional tubelet projection, 

(2) a \textbf{Multi-Head Self-Attention (MHSA) Encoder}, and 

(3) a \textbf{Linear Patch Decoder}. 

The framework exploits local temporal context preservation within MSI 
time-series to reconstruct spectral information in cloud-covered pixels,  while integrating unmasked SAR observations that penetrate cloud cover as complementary all-weather features for multi-modal reconstruction.

\subsubsection{Temporal-Spatial Fusion Embedding via 3D Convolutional Tubelet Projection:}\label{sec:TSFE}


The key innovation is the introduction of a constrained temporal span in the tubelet extraction process. SMTS-ViT typically concatenates the entire time-series channels with spatial channels (i.e., temporal span $t = T$), creating a channel dimension $C' = T \times C$. This full-sequence aggregation approach leads to excessive cross-day information mixing and substantial information loss.

In contrast, our proposed temporal-spatial fusion embedding extracts non-overlapping spatio-temporal tubelets using 3D convolution with a short, fixed temporal span $(t = 2)$. This design principle reduces information degradation compared to full-sequence aggregation: as $t \to T$, transient spectral variations critical for accurate cloud reconstruction become obscured within the aggregated channels.





The temporal-spatial fusion embedding operates on the input tensor directly via 3D convolutional filtering. Unlike 2D convolution applied independently to each timeframe, 3D convolution simultaneously processes spatial and temporal dimensions, enabling implicit modeling of local spatio-temporal patterns within the input volume.

For the MSI-only scenario, the input tensor $X_{\text{MSI}} \in \mathbb{R}^{T \times C_{\text{MSI}} \times H \times W}$ is transformed into a sequence of non-overlapping spatio-temporal tubelets. For the SAR-MSI fusion scenario, the concatenated input tensor $X \in \mathbb{R}^{T \times (C_{\text{MSI}} + C_{\text{SAR}}) \times H \times W}$ combines all spectral and polarization channels. The 3D convolution operates with kernel size $(k_t, k_s, k_s)$, where the temporal kernel dimension is $k_t = t$ (the constrained temporal span, fixed at $t=2$ in our implementation), and the spatial kernel dimension is $k_s = 5$ (producing $5 \times 5$ pixel patches).

The 3D convolutional operation for extracting the $p$-th tubelet is defined as:

\begin{equation}\label{equ:Tubelet3D}
X'_{(p, t')} = \sum_{t''=0}^{k_t-1} \sum_{i=0}^{k_s-1} \sum_{j=0}^{k_s-1} \sum_{c=1}^{C} W_{(p, c, i, j, t'')} X_{(t'+t'', c, h+i, w+j)} + b_p
\end{equation}

\begin{tabbing}
where $\qquad$ \= $X'_{(p, t')}$ = embedding feature for the $p$-th tubelet  \\
\> $\qquad$ at temporal position $t'$\\
\> $p$ = tubelet index \\
\> $t'$ = temporal anchor position of tubelet\\
\> $t''$ = temporal offset within the convolution kernel\\
\> $i, j$ = spatial offsets within the convolution kernel\\
\> $X_{(t'+t'', c, h+i, w+j)}$ = input pixel value at temporal\\
\> $\qquad$  position $(t' + t'')$\\
\> $W_{(p, c, i, j, t'')}$ = 3D convolution kernel weight\\
\> $b_p$ = bias term for patch \\

\end{tabbing}

This produces a sequence of patch tokens, each jointly encoding spatial structure and local temporal evolution. The linear embedding layer then projects each tubelet into the embedding space, following by the positional encoding, with dimension $d_e$:

\begin{equation}\label{equ:LinearEmbed}
T_{(p)} = \text{Linear}(X'_{(p, t')}) \in \mathbb{R}^{d_e}
\end{equation}

\begin{tabbing}
where \hspace{0.8cm} \= $T_{(p)}$ = $p$-th tubelet token embedding\\
\> $d_e$ = embedding dimension\\
\> $\text{Linear}(\cdot)$ = learned linear transformation layer
\end{tabbing}








\subsubsection{Multi-Head Self-Attention Encoder with Joint Temporal-Spatial Attention:}\label{sec:MHSA}

The MHSA mechanism operates directly on the sequence of spatial-temporal patch (tubelet) embeddings produced by 3D convolution, allowing the model to capture complex dependencies across both temporal and spatial dimensions in the data. Before entering the MHSA blocks, each tubelet embedding is summed with its corresponding positional encoding, ensuring the network is aware of the patch's location within both space and time. This joint encoding strategy enables the model to learn interactions between different moments and locations within the input sequence in a unified way.

In each self-attention block, for each attention head, query ($Q$), key ($K$), and value ($V$) matrices are computed from the input tubelet embeddings:

\begin{equation}\label{equ:Attention}
\text{Attention}(Q, K, V) = \text{Softmax}\left(\frac{QK^T}{\sqrt{d_k}}\right)V
\end{equation}

\begin{tabbing}
where $\qquad$ \= $Q$ = query matrix, dimension: $N_{\text{patch}} \times d_k$\\
\> $K$ = key matrix, dimension: $N_{\text{patch}} \times d_k$ \\
\> $V$ = value matrix, dimension: $N_{\text{patch}} \times d_v$ \\
\> $d_k$ = dimension of key vectors \\

\end{tabbing}

Multi-head attention employs $h$ parallel attention heads to process the input, which are then concatenated and projected back to the embedding dimension:

\begin{equation}\label{equ:MultiHead}
\text{MultiHead}(Q, K, V) = \text{Concat}(\text{head}_1, \ldots, \text{head}_h)W^O
\end{equation}

\begin{tabbing}
where $\qquad$ \= $h$ = number of attention heads\\
\> $W^O$ = output projection matrix,\\
\> $\qquad$ dimension: $h \cdot d_v \times d_e$
\end{tabbing}

In our framework, joint temporal-spatial attention is implemented by combining positional encoding with all tubelet (patch) embeddings and processing them through Multi-Head Self-Attention blocks. This structure enables the model to automatically discover and emphasize relevant relationships across both time and space, without requiring explicit masking. As a result, the attention mechanism flexibly aggregates long-range temporal and spatial features that are informative for reconstructing MSI data under cloud cover and synergistically integrating complementary SAR signals. 

\subsubsection{Linear Patch Decoder:}

The decoder consists of a linear projection layer that reshapes high-dimensional encoder features back into the original input time-series MSI structure. The encoder output feature maps are projected linearly to reconstruct patches corresponding to the input spatial-temporal resolution:

\begin{equation}\label{equ:Decoder}
\hat{Y} \in \mathbb{R}^{T \times C_{\text{MSI}} \times H \times W}
\end{equation}

\begin{tabbing}
where \hspace{0.8cm} \= $\hat{Y}$ = reconstructed MSI tensor,\\
\> dimensions match the input tensor $X_{\text{MSI}}$
\end{tabbing}

This linear reconstruction design is chosen for its simplicity and efficiency, enabling the model to learn direct mappings from learned representations to pixel-level reconstructions.

\subsection{Multi-Scale Loss Function:}

A multi-scale loss function is employed for the training process that combines two complementary objectives: Mean Squared Error (MSE) loss for pixel-level reconstruction accuracy, and Spectral Angle Mapper (SAM) loss for spectral consistency. The multi-scale formulation ensures reconstruction quality across multiple spatial resolutions:

\begin{equation}\label{equ:MultiScaleLoss}
\mathcal{L}_{\text{Multi-Scale}} = \sum_{s=1}^{S} w_{\text{MSE}}^{(s)} \cdot \mathcal{L}_{\text{MSE}}^{(s)} + w_{\text{SAM}}^{(s)} \cdot \mathcal{L}_{\text{SAM}}^{(s)}
\end{equation}

\begin{tabbing}
where \hspace{0.8cm} \= $S$ = number of pyramid scales\\
\> $w_{\text{MSE}}^{(s)}$ = weight for MSE loss at scale $s$\\
\> $w_{\text{SAM}}^{(s)}$ = weight for SAM loss at scale $s$\\
\> $s$ = scale index
\end{tabbing}

Images are rescaled using bilinear interpolation to generate the multi-scale pyramid. 

\textbf{MSE Loss} measures global pixel-level fidelity:

\begin{equation}\label{equ:MSELoss}
\begin{split}
\mathcal{L}_{\text{MSE}} &= \frac{1}{N \cdot T \cdot C \cdot H \cdot W} \\
&\quad \times \sum_{n=1}^N \sum_{t=1}^T \sum_{c=1}^C \sum_{h=1}^H \sum_{w=1}^W \left(y_{\text{true}}^{(n,t,c,h,w)} - y_{\text{pred}}^{(n,t,c,h,w)}\right)^2
\end{split}
\end{equation}

\begin{tabbing}
where \hspace{0.8cm} \= $N$ = batch size\\
\> $y_{\text{true}}^{(n,t,c,h,w)}$ = true MSI image pixel spectral value \\
\> $y_{\text{pred}}^{(n,t,c,h,w)}$ = predicted pixel spectral value\\
\end{tabbing}

\textbf{SAM Loss} enforces spectral consistency by computing the angular distance between true and predicted spectral vectors at each pixel:


\begin{equation}\label{equ:SAMLoss_split}
\begin{split}
\mathcal{L}_{\text{SAM}} &= \frac{1}{N \cdot T \cdot H \cdot W} \sum_{n=1}^N \sum_{t=1}^T \sum_{(h,w)=1}^{H \times W} \\
&\quad \arccos\left(\frac{\langle \mathbf{y}_{\text{true},h,w}^{(n,t)}, \mathbf{y}_{\text{pred},h,w}^{(n,t)} \rangle}{\|\mathbf{y}_{\text{true},h,w}^{(n,t)}\|_2 \cdot \|\mathbf{y}_{\text{pred},h,w}^{(n,t)}\|_2 + \epsilon}\right)
\end{split}
\end{equation}

\begin{tabbing}
where \hspace{0.8cm} \= $\mathbf{y}_{\text{true},h,w}^{(n,t)}$ = spectral vector at pixel $(h,w)$ and time $t$\\
\> $\mathbf{y}_{\text{pred},h,w}^{(n,t)}$ = predicted spectral vector \\
\>  \hspace{1.4cm} at pixel $(h,w)$ and time $t$\\
\> $\langle \cdot, \cdot \rangle$ = dot product\\
\> $\|\cdot\|_2$ = Euclidean norm\\
\> $\epsilon$ = small constant to prevent numerical instability\\
\end{tabbing}










\section{Experiments}\label{Exp}
\begin{figure}[ht!]
\begin{center}
		\includegraphics[width=1\columnwidth]{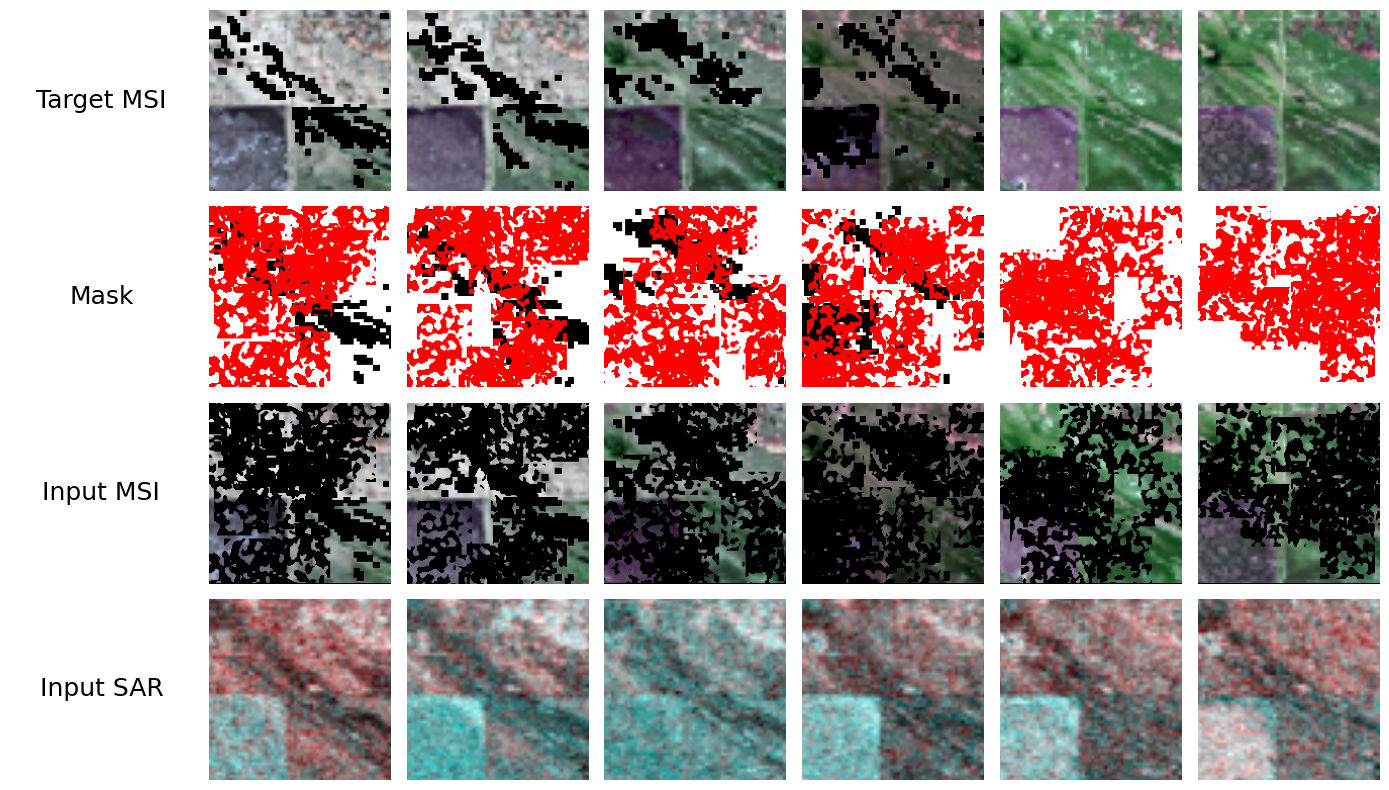}
	\caption{Overview of Multi-Modal Data Inputs and Cloud Masks. There are four rows: the first row shows target MSI reconstruction images, the second displays cloud masks (where black pixels indicate real cloud occlusion and red pixels indicate artificial cloud coverage), the third row contains input MSI data, and the fourth row presents input SAR data for the ViViT model. The x-axis indicates the timeline of data acquisition.}
\label{fig:setup}
\end{center}
\end{figure}

\begin{table*}[h]
\centering
\begin{tabular}{|l|c|c|c|c|c|}\hline
\textbf{Model} & \textbf{\# Cloud} & \textbf{MSE ($\times 10^{-3}$)} $\downarrow$ & \textbf{SAM($\times 10^{-1}$)} $\downarrow$ & \textbf{PSNR} $\uparrow$ & \textbf{SSIM} $\uparrow$ \\ \hline

\multirow{2}{*}{MTS-ViT}
& 20 & 4.435 & 1.036 & 23.293 & 0.796 \\
& 30 & 7.131 & 1.402 & 22.256 & 0.727 \\ \hline

\multirow{2}{*}{MTS-ViViT}
& 20 & 4.336 & 0.966 & 23.574 & 0.814 \\
& 30 & 5.002 & 1.131 & 22.925 & 0.772 \\ \hline

\multirow{2}{*}{SMTS-ViT}
& 20 & \underline{3.464} & \underline{0.933} & \underline{25.242} & \underline{0.849} \\
& 30 & 3.966 & 1.071 & 24.052 & 0.783 \\ \hline

\multirow{2}{*}{SMTS-ViViT}
& 20 & \textbf{3.106} & \textbf{0.857} & \textbf{25.649} & \textbf{0.867} \\
& 30 & 3.784 & 1.017 & 24.735 & 0.813 \\ \hline

MTS-ViT (AVG) & - & 5.783 & 1.219 & 22.775 & 0.762 \\
MTS-ViViT (AVG) & - & 4.669 & 1.049 & 23.250 & 0.793 \\ 

SMTS-ViT (AVG) & - & \underline{3.715} & \underline{1.002} & \underline{24.647} & \underline{0.816} \\
SMTS-ViViT (AVG) & - & \textbf{3.445} & \textbf{0.937} & \textbf{25.192} & \textbf{0.840} \\ \hline

\end{tabular}
\caption{Comprehensive evaluation results across all model configurations and cloud count scenarios. MTS-ViT and MTS-ViViT represent MSI-only reconstruction, while SMTS-ViT and SMTS-ViViT represent SAR-MSI fusion. Cloud counts (20, 30) represent increasing occlusion severity. Arrows indicate optimization direction: $\downarrow$ (lower better), $\uparrow$ (higher better). Best results highlighted in bold. The second-best results are underlined.}
\label{tab:Comprehensive_Results}
\end{table*}

\subsection{Experimental Setup}\label{sec:ExpSetup}

As shown in Figure \ref{fig:setup}, for the experimental setup, the MSI-only scenario uses a sequence of MSI as input, while the MSI and SAR fusion scenario uses both time-series MSI and co-registered SAR images as input. In both cases, a cloud mask is provided for each image, consisting of two parts: the original cloud mask and an additional artificial mask to simulate more cloud-covered areas. The output is the reconstructed time-series cloud-free MSI images. Four model variants were evaluated to assess the contribution of time-spatial fusion embedding:

\begin{itemize}
\item \textbf{MTS-ViT \cite{li2025vision}:} Baseline multi-temporal Vision Transformer with standard 2D convolutional patch projection using full-sequence temporal aggregation of MSI ($t = T = 6$, $C_{\text{total}} = C_{\text{MSI}} = 11$).

\item \textbf{MTS-ViViT:} Proposed multi-temporal Video Vision Transformer with 3D convolutional tubelet extraction using constrained temporal span ($t = 2$, $C_{\text{total}} = C_{\text{MSI}} = 11$).

\item \textbf{SMTS-ViT \cite{li2025vision}:} Baseline SAR-MSI fusion with standard temporal aggregation of both MSI and SAR ($t = T$, $C_{\text{total}} = C_{\text{MSI}} + C_{\text{SAR}} = 13$).

\item \textbf{SMTS-ViViT:} Proposed SAR-MSI fusion with 3D tubelet projection ($t = 2$, $C_{\text{total}} = C_{\text{MSI}} + C_{\text{SAR}} = 13$).
\end{itemize}

\subsection{Datasets and Settings}\label{sec:DatasetsSettings}


The dataset was partitioned into 80\% for training and 20\% for validation with spatial stratification. Training and validation data were subdivided into $60 \times 60$ pixel patches. To simulate realistic cloud occlusion, artificial cloud coverage was introduced during training using Gaussian smooth random noise with configurable cloud size (default $\text{cloud\_size} = 0.3$, i.e., clouds initiate at 30\% of image dimensions) and cloud count (default $\# \text{Clouds} = 20$ or $30$). Cloud masks were applied via element-wise multiplication to corrupt MSI data while preserving SAR observations. 


\subsubsection{Hyperparameter Configuration:} All models share \\ identical hyperparameter settings: patch size $5 \times 5$ pixels, network depth 6 encoder layers, 8 multi-head self-attention heads, embedding dimension $d_e = 64$, and feed-forward hidden dimension $4 \times d_e = 256$. Training is conducted for 200 epochs with batch size 8 using Adam optimizer ($\beta_1 = 0.9, \beta_2 = 0.999$) and learning rate $\eta_0 = 1 \times 10^{-3}$. Learning rate decay is applied at rate $\gamma = 0.95$ every 10 epochs. Random seed is fixed at 42 for reproducibility.

Loss function combines MSE and SAM with equal weights: $w_{\text{MSE}} = w_{\text{SAM}} = 0.5$. Multi-scale pyramid consists of $S = 3$ scales at resolutions $\{1.0, 0.5, 0.25\}$. The temporal span for baseline models is $t_{\text{ViT}} = T = 6$ time steps, while proposed models employ $t_{\text{ViViT}} = 2$ time steps. The 3D convolution kernel size is $(2, 5, 5)$ for temporal and spatial dimensions.

\subsubsection{Robustness Assessment}

Model robustness is trained and evaluated across multiple cloud count configurations:  $\# \text{Clouds} $ = $20$ and $30$. Each cloud configuration represents increasingly severe occlusion scenarios. For each configuration, a minimum of three independent runs are conducted using different random initializations, and results are averaged for statistical assessment.

\subsubsection{Evaluation Metrics}

Performance is quantified using four complementary metrics: 
\\
(1) MSE measuring pixel-level reconstruction error,\\
(2) SAM measuring angular deviation in spectral space,\\
(3) Peak Signal-to-Noise Ratio (PSNR) evaluating signal quality in decibels, computed as 

\begin{equation}\label{equ:psnr_def_10log}
\mathrm{PSNR}( X_{\text{MSI}},\hat Y) = 10\log_{10}\left(\frac{\mathrm{MAX}_{X_{\text{MSI}}}^{2}}{\mathrm{MSE}(X_{\text{MSI}},\hat Y)}\right)
\end{equation}
\begin{tabbing}
where \hspace{0.8cm} \= \text{MAX}$_{X_{\text{MSI}}} = 1$, image dynamic range \\
\>  \hspace{1.4cm} for normalized data\\
\end{tabbing}
(4) Structural Similarity Index Measure (SSIM) assessing perceptual quality in range $[-1, 1]$:
\begin{equation}\label{equ:ssim_def}
\mathrm{SSIM}(x,y)=\frac{(2\mu_x\mu_y + C_1)(2\sigma_{xy} + C_2)}{(\mu_x^{2} + \mu_y^{2} + C_1)(\sigma_x^{2} + \sigma_y^{2} + C_2)}
\end{equation}

\begin{tabbing}
where \hspace{0.6cm} \= 
$\mu_x,\,\mu_y$ = local means of $x$ and $y$ \\
\> $\sigma_x^{2},\,\sigma_y^{2}$ = local variances of $x$ and $y$ \\
\> $\sigma_{xy}$ = local covariance between $x$ and $y$ \\
\> $C_1=(K_1L)^2$, $C_2=(K_2L)^2$, stabilization constants\\
\> $L= 1$, image dynamic range for normalized data \\
\> $K_1,\,K_2$ = small positive constants \\
\>  \hspace{1.0cm} (typically $K_1=0.01,\,K_2=0.03$) \\
\end{tabbing}

\subsection{Results}\label{sec:Results}
Table \ref{tab:Comprehensive_Results} summarizes the quantitative performance of all models under varying cloud conditions, using four standard metrics: MSE ($\times 10^{-3}$), SAM ($\times 10^{-1}$), PSNR, and SSIM. Lower MSE and SAM scores, alongside higher PSNR and SSIM values, indicate better image reconstruction quality.

Across all groups, the proposed SMTS-ViViT consistently achieves the best average performance, with the lowest average MSE (3.445), lowest SAM (0.937), highest PSNR (25.192), and highest SSIM (0.840). SMTS-ViT ranks as the second-best approach, outperforming both baseline methods (MTS-ViT and MTS-ViViT) on all metrics. Under moderate cloud levels (\#Clouds = 20), SMTS-ViViT and SMTS-ViT demonstrate clear improvements over their standard counterparts, with SMTS-ViViT yielding the lowest errors and highest perceptual scores.

Importantly, when the cloud amount increases to 30, reconstruction quality decreases for all models, as expected. This demonstrates the robustness of the temporal-spatial embedding approach, especially when combined with the ViViT backbone, for reconstructing multispectral time series data under heavy cloud occlusions.

Comparing integration of SAR and MSI vs. MSI-only:
Adding SAR auxiliary information substantially improves reconstruction performance. For \#Clouds = 20:

Comparing MTS-ViT (MSI only) and SMTS-ViT (SAR+MSI), the inclusion of SAR reduces MSE by 21.98\%, reduces SAM by 9.94\%, increases PSNR by 8.37\%, and increases SSIM by 6.66\%.

Comparing MTS-ViViT and SMTS-ViViT, adding SAR decreases MSE by 28.37\%, decreases SAM by 11.28\%, increases PSNR by 8.09\%, and increases SSIM by 6.51\%.

These improvements are consistent with prior work showing that fusing SAR with optical data can significantly enhance cloud removal and image reconstruction performance. SAR provides robust complementary information when optical bands are corrupted by clouds, resulting in consistently better quantitative and perceptual results.

ViViT backbone benefits:
For cloud=20, integrating the ViViT transformer backbone yields additional gains:

MTS-ViT vs. MTS-ViViT: ViViT reduces MSE by 2.23\%.

SMTS-ViT vs. SMTS-ViViT: ViViT reduces MSE by 10.33\%.

Similar improvements are observed under heavier cloud cover (\#Clouds = 30), reinforcing the advantage of transformer-based architectures for time-series remote sensing tasks.

In summary, these results validate that fusing SAR data with MSI significantly boosts cloud removal accuracy over MSI-only baselines. The combination of our sparse-in-time modeling strategy with a transformer backbone (ViViT) yields state-of-the-art performance on all metrics and under all cloud conditions.

\subsection{Visualization}\label{sec:Visual}

\begin{figure}[ht!]
\begin{center}
		\includegraphics[width=1\columnwidth]{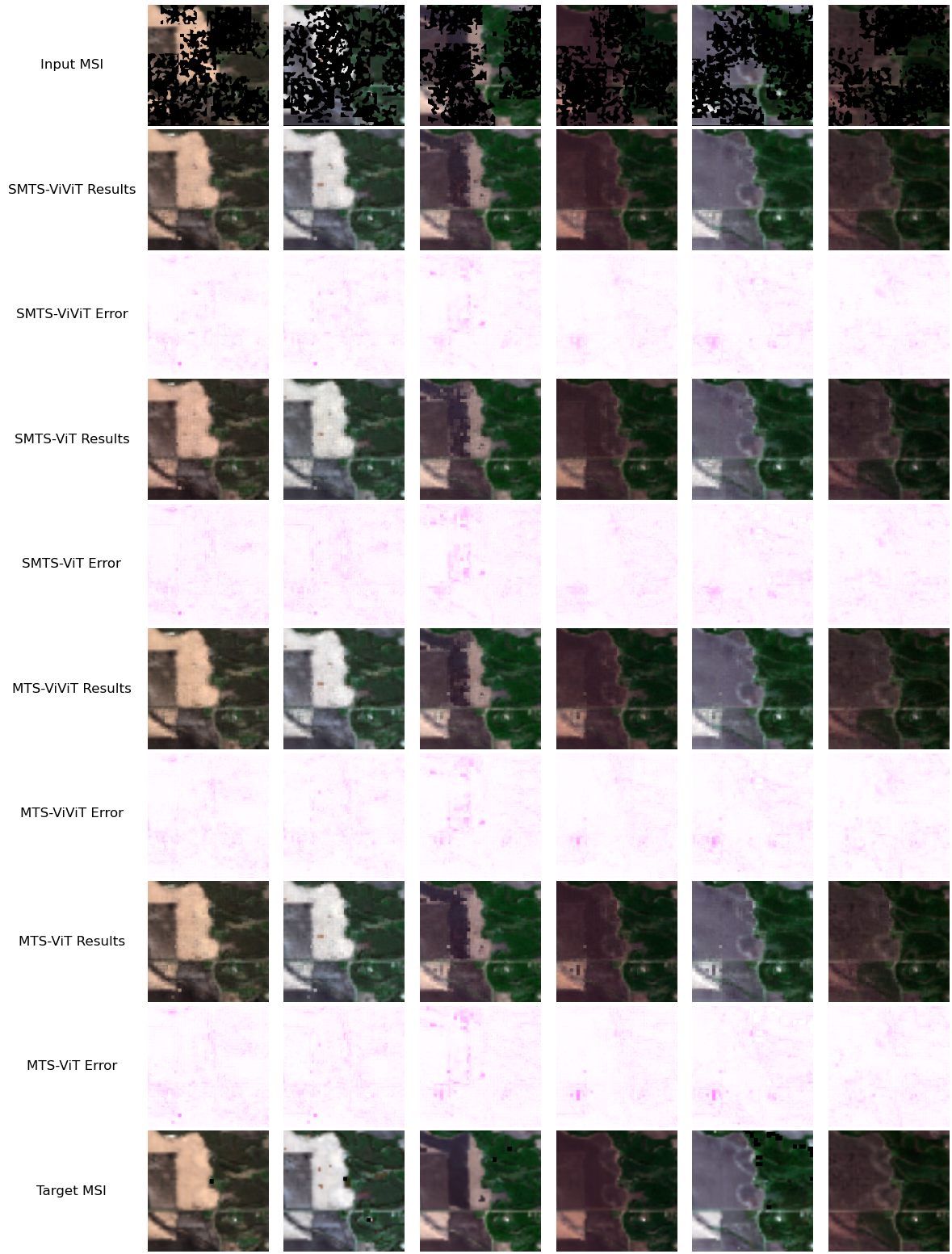}
	\caption{Cloud Removal and Reconstruction Results Across Models. Black pixels present cloud masks.}
\label{fig:visual}
\end{center}
\end{figure}

To qualitatively assess model performance, a comprehensive visualization is presented in Figure \ref{fig:visual}. The figure displays ten rows: the input MSI image (with cloud mask), the reconstruction and error maps for SMTS-ViViT, SMTS-ViT, MTS-ViViT, and MTS-ViT, as well as the clear target MSI as a reference. The error maps are shown to highlight the deviations between each model's prediction and the ground truth.

From these visual comparisons, several trends are evident. SMTS-ViViT consistently produces reconstructions with the least visible residual error, closely matching the target MSI even in heavily clouded regions. As seen in the third row of the figure, the pink color of the error portion in the SMTS-ViViT result is lighter than the error hue in other results, indicating smaller errors. Furthermore, the errors in the time series images 1 and 3 also show significantly less error in the details.

Both SMTS-ViViT, SMTS-ViT recover more spatial detail and preserve spectral characteristics better than their counterparts using only MSI input. Error maps for SMTS-based models exhibit fewer and less intense artifacts.

A clearer distinction between the SMTS- and MTS-based reconstructions, as well as between the ViViT- and ViT-based reconstructions, is evident not only quantitatively but also visually. These qualitative differences further confirm the added value of fusing SAR information and highlight the effectiveness of the proposed temporal-spatial embedding approach for improved multispectral cloud removal and reconstruction.

\section{Conclusions}\label{Conclusions}

This work proposes a ViViT-based model with temporal-spatial time embedding for cloud removal by fusing MSI and SAR, addressing the challenge of accurate time-series MSI reconstruction under different cloud conditions. Our main contribution is a temporal-spatial time embedding mechanism that allows the model to effectively leverage observations from multiple dates by explicitly encoding temporal information in both space and time, which leads to significantly better results than MSI-only or simple fusion approaches.

Our experiments show that combining SAR and MSI significantly boosts performance on all tested metrics, and the sparsity-aware temporal embedding further improves robustness under heavy clouds. The SMTS-ViViT model, which integrates this embedding with a transformer backbone, achieves the best results both quantitatively and visually.

However, the current version still aggregates MSI and SAR for each pixel as a mean feature across dates. For future work, we plan to feed all single-date MSI and SAR data separately and develop time embedding strategies that can flexibly accommodate different observation dates in different years or regions. Such methods will allow the network to understand and utilize information even when the set of available dates varies, opening new possibilities for robust cross-temporal, cross-regional analysis and applications in remote sensing.

{
	\begin{spacing}{1.17}
		\normalsize
		\bibliography{ISPRSguidelines_authors} 
	\end{spacing}
}

\end{document}